\documentclass{article}

\usepackage[preprint, nonatbib]{neurips_2020}

\usepackage[utf8]{inputenc} 
\usepackage[T1]{fontenc}    
\usepackage{hyperref}       
\usepackage{url}            
\usepackage{booktabs}       
\usepackage{amsfonts}       
\usepackage{nicefrac}       
\usepackage{microtype}      

\title{Generative Adversarial Training Can Improve Neural Language Models}

\author{
  Sajad Movahedi \\
  University of Tehran\\
  \texttt{s.movahedi@ut.ac.ir} \\
  \And 
  Azadeh Shakery \\
  University of Tehran \\
  \texttt{shakery@ut.ac.ir}\\
}

\begin{document}

\maketitle

\begin{abstract}
  While deep learning in the form of recurrent neural networks (RNNs) has caused a significant improvement in neural language modeling, the fact that they are extremely prone to overfitting is still a mainly unresolved issue. In this paper we propose a regularization method based on generative adversarial networks (GANs) and adversarial training (AT), that can prevent overfitting in neural language models. Unlike common adversarial training methods such as the fast gradient sign method (FGSM) that require a second back-propagation through time, and therefore effectively require at least twice the amount of time for regular training, the overhead of our method does not exceed more than 20\% of the training of the baselines.
  
\end{abstract}

\section{Introduction}
Language modeling is among the most fundamental tasks in natural language processing (NLP), with one of the major improvements in recent years in NLP coming from utilizing neural language models for inductive transfer learning \cite{devlin2018bert, yang2019xlnet, liu2019roberta}. With the introduction of long-short term memory networks (LSTMs), the task of language modeling has seen a significant improvement in performance \cite{merity2017regularizing}, albeit at the cost of longer training time \cite{merity2018analysis}.
While extremely effective LSTMs suffer from severe over-fitting, which has been addressed mainly through dropout methods \cite{NIPS2016_6241, zolna2017fraternal}, and yet there still exists a large gap between the training and test performance.
\par A word-level language model aims to estimate the probability of any given string of words $x_{1:T}=[x_1, x_2, ..., x_T]$. From the chain rule, we can write: 
$$p(x_{1:T})=p(x_1, x_2, ..., x_T)=\Pi_{t=1}^{T}p(x_t|x_{1:t-1})$$
Given the recurrent structure of recurrent neural networks (RNNs), we can estimate the conditional probability $p(x_t|x_{1:t-1})$ by using the hidden state of the network at the $t^{th}$ time-step. Let $h_t \in R^{dim(h)}$ be the $t^{th}$ output of the RNN. So we have 
$h_t = f(x_{t-1}, h_{t-1}; \theta)$, where $x_{t-1}$ is a vector representation of the $t-1^{th}$ word in the sequence, $h_{t-1}$ is the $t-1^{th}$ hidden representation of the RNN, and $\theta$ represents the parameters of the RNN. Using the softmax function, we can represent a probability distribution over the vocabulary:
$$p(x_t|x_{1:t-1}, \theta)=Softmax(h_t, \omega)$$
Where $\omega \in R^{dim(h) \times dim(vocab)}$ represents the weights of the final layer of the network.

\par One of the most effective methods for improving the robustness of deep learning models towards adversarial data is adversarial training \cite{kurakin2016adversarial}, which involves adding perturbations with a controlled magnitude to the training data. Recently, it has been observed that adversarial training may have a positive effect over the performance of text classification \cite{miyato2016adversarial}, image classification \cite{xie2019adversarial}, and sequence labeling \cite{yasunaga2017robust}, which is mainly attributed to preventing over-fitting and improving generalization \cite{xie2019adversarial}.
\par The adversarial training framework aims to decrease the worst-case loss of adversarial data \cite{kurakin2016adversarial}. We can write this objective in the following min-max form:
$$\min_\theta \max_{\Vert r \Vert_2 \leq \epsilon} \mathcal{L}(f(x + r; \theta), y)$$
Where $\mathcal{L}$ is the loss function, which is usually defined as negative log likelihood, $y$ is the observed output, and $r$ is the adversarial perturbation added to the input, the norm of which is limited by $\epsilon$. The most common way of estimating $r$ is a method called the fast gradient method \cite{miyato2016adversarial}, which utilizes the gradient of the model with respect to input as a first-order approximation of $r$:
\begin{equation}
    r = \epsilon \frac{\nabla_x f(x, \theta)}{\Vert \nabla_x f(x, \theta) \Vert_2}
\end{equation}
Estimating $r$ requires another back-propagation, and effectively doubles the training time in adversarial training, which has been the subject of some research recently \cite{shafahi2019adversarial, zhang2019you}. 
\par Here we propose a method to prevent over-fitting in neural language models through adversarial training. Instead of using the gradients of the model to estimate the adversarial data, we propose to use the generative adversarial network (GAN) framework to produce the perturbation while training the adversarial network simultaneously. Using our method, we were able to decrease the over-head of adversarial training to less than 20\% of the training time of the baseline models, while also improving the performance of the baseline.

\section{Method}
We propose to estimate the adversarial perturbation corresponding to the $t^{th}$ word $r_t$ by using another RNN as a generator network. Let $g(.)$ denote such network, with the set of weights $\eta$. Then we have $r_t = g(x_{t-1}, h'_{t-1}, \eta)$, where $h_{t-1}'$ is the current hidden state of the adversarial RNN. In order to estimate $\eta$, we propose to train the neural language model and the adversarial model jointly, with $\theta$ updated using gradient descent (to minimize $\mathcal{L}(f(x+r, \theta), y)$ over $\theta$) and $\eta$ updated using gradient ascent (to maximize $\mathcal{L}(f(x+g(x, \eta), \theta), y)$ over $\eta$) in a single mini-batch. We can also control the magnitude of $r_t$, by adding the following term to the loss function:
$$ \mathcal{L}_{reg} = \frac{1}{T} \sum_{t=1}^{T}\max(0, \Vert r_t \Vert_2 - \alpha \times \Vert x_t \Vert_2)$$
In this case, for the $t^{th}$ word we have $\epsilon_t = \alpha \Vert x_t \Vert_2$.

\par In our preliminary experiments, we noticed that in order to prevent the adversarial network from overfitting, we need to use dropout, which will decrease the effectiveness of the perturbation vectors $r$. In this case, we can use the Monte Carlo method to find a better perturbation vector, using the output of $g(x_t, h'_{t-1}, \eta)$ with different dropout masks as random samples of the distribution of the optimal value of $r_t$. So we define our estimation as the following iterative method:
$$\hat{r}_t^k=\sum_{i=0}^{k-1} \frac{1}{k} g(x_t + \hat{r}_t^{i}, (h'_{t-1})^i, \sigma_k \odot \eta)$$
Where $\sigma_k$ is the $k^{th}$ random dropout mask. Note that because of the iterative structure, this definition isn't exactly in line with the Monte Carlo method. The main reason behind this difference is that the iterative structure performed better than averaging $\hat{r}_t^k$s in our preliminary experiments. Also note that this structure is very similar to the projected gradient descent method \cite{kurakin2016adversarial}, which performs first order approximation using fast gradient method multiple times. Furthermore, in order to train the adversarial network simultaneously, we only need to back-propagate from the first iteration $\hat{r}_t^1$, and we can treat $\eta$ in the rest of the iterations as constant, which will greatly improve the training speed. We also noticed that random starts \cite{madry2017towards} can help both the effectiveness of our adversarial method and the improvement in the task. So we set $\hat{r}_t^0=\epsilon_0 \times \mathcal{N}(0, I)$, where $\mathcal{N}(0, I)$ is a random multi-variate standard normal distribution, and $\epsilon_0$ is a scaling factor.

\par  The overhead of our method is at most as much as the overhead of back-propagation over the weights of the adversarial model, which we can control by using simpler models. In our experiments, we found the best model adversarial to be a single layer LSTM. So the overall overhead was around 20\% of training without adversarial perturbations.
\section{Experiments}
We performed experiments on the PTB corpus \cite{marcus1993building} and WikiText2 \cite{merity2016pointer}. Our baseline was the AWD-LSTM method \cite{merity2017regularizing}, which consists of three LSTM cells stacked on top of each other, and a plethora of regularization methods to decrease overfitting. Note that the training procedure of AWD-LSTM involves a training stage and a fine-tuning stage after that, but we only report the finetuning stage results due to lack of space. We also report the performance of some of the earlier works. The results of the PTB dataset and the WikiText2 dataset are presented in \tableautorefname{1}.

\begin{table}[h]
    \scriptsize
    \centering
    \begin{tabular}{c|c|c|c|c|c}
    \multicolumn{3}{c|}{PTB} &\multicolumn{3}{|c}{WikiText2} \\\hline
    Method & Valid & Test &  Method & Valid & Test\\ \hline
    Variational LSTM + weight tying \cite{inan2016tying} & 71.1 & 68.5 & Variational LSTM + weight tying \cite{inan2016tying} & 91.5 & 87.0  \\
    NAS-RNN \cite{zoph2016neural} & - & 62.4 &DARTS \cite{liu2018darts} & 69.5 & 66.9 \\
    DARTS \cite{liu2018darts} & 58.3 & 56.1 & & & \\ \hline
    AWD-LSTM \cite{merity2017regularizing} & 60.00 & 57.30 & AWD-LSTM \cite{merity2017regularizing} & 68.60 & 65.80  \\
    AWD-LSTM + Adversarial Training & \textbf{57.61} & \textbf{55.46} & AWD-LSTM + Adversarial Training & \textbf{65.39} & \textbf{62.65}\\
    \end{tabular}
    \caption{Perplexity of the validation set and the test set for the WikiText2 dataset. Our method is AWD-LSTM + Adversarial Training. Smaller perplexity means better performance.}
    \label{tab:my_label}
\end{table}
\bibliographystyle{plain} 
\bibliography{library.bib}

\end{document}